\documentclass[runningheads]{llncs}
\usepackage[T1]{fontenc}
\usepackage{graphicx}
\usepackage{hyperref}
\usepackage{xcolor}
\hypersetup{
    colorlinks,
    linkcolor={blue!50!black},
    citecolor={blue!50!black},
    urlcolor={blue!50!black}
}
\usepackage{color}
\usepackage{bbm}
\usepackage{amsmath}
\usepackage{amsfonts}
\usepackage{multirow}
\usepackage{enumitem}
\begin{document}
\title{Local Attention Graph-based Transformer for Multi-target Genetic Alteration Prediction}
\titlerunning{Local Attention Graph-based Transformer for Genetic Alteration Prediction}
%

\author{Daniel Reisenb\"uchler\inst{1,2} \and
Sophia J. Wagner\inst{1,2} \and
Melanie Boxberg\inst{4} \and
Tingying Peng\inst{2}}

\authorrunning{D. Reisenb\"uchler et al.}
%
\institute{Technical University Munich, Munich, Germany \and
Helmholtz AI, Neuherberg, Germany \and
Institute of Pathology Munich-North, Munich, Germany \newline reisenbuechler@helmholtz-muenchen.de
}

\maketitle 
\begin{abstract} 

Classical multiple instance learning (MIL) methods are often based on the identical and independent distributed assumption between instances, hence neglecting the potentially rich contextual information beyond individual entities. On the other hand, Transformers with global self-attention modules have been proposed to model the interdependencies among all instances. However, in this paper we question: Is global relation modeling using self-attention necessary, or can we appropriately restrict self-attention calculations to local regimes in large-scale whole slide images (WSIs)? 
We propose a general-purpose local attention graph-based Transformer for MIL (LA-MIL), introducing an inductive bias by explicitly contextualizing instances in adaptive local regimes of arbitrary size. Additionally, an efficiently adapted loss function enables our approach to learn expressive WSI embeddings for the joint analysis of multiple biomarkers. We demonstrate that LA-MIL achieves state-of-the-art results in mutation prediction for gastrointestinal cancer, outperforming existing models on important biomarkers such as microsatellite instability for colorectal cancer. Our findings suggest that local self-attention sufficiently models dependencies on par with global modules. Our LA-MIL implementation is available at \href{https://github.com/agentdr1/LA_MIL}{https://github.com/agentdr1/LA\_MIL}.

\end{abstract}

\section{Introduction}

Advances in slide-scanning microscopes and deep learning-based image analysis have significantly increased interest in computational pathology~\cite{Cooper2018}. Whole slide images typically contain billions of pixels and reach up to several gigabytes in size. To mitigate the resulting computational burden, WSIs are commonly tessellated into smaller tiles~\cite{Murchan2021}. However, patient diagnosis is typically only available as weakly-supervised slide-level annotation, e.g., cancer vs.\ non-cancer classification, cancer subtyping, or genomic analyses. 

In histopathological image analysis, this task is formulated as multiple instance learning (MIL), where a WSI is considered as a bag, and tiles as contained instances. Hence, efficiently learning representations and aggregating them from tiles to a bag label is crucial. One simple solution for this is to pass the bag label onto each tile, reducing MIL to supervised learning. This approach is particularly favored because of its ease of implementation, e.g., to predict microsatellite instability or tumor mutational burden~\cite{Kather_msi,Wang2020}. The final bag-level prediction is obtained by aggregating all instance-level predictions with average pooling. These methods have two drawbacks: (i) a fraction of instance labels may differ from the bag label and therefore form label noise in supervised learning, and (ii) no morphological or spatial correlation between tiles is taken into account. 

To remedy (i), MIL can learn from bag-level annotation without assuming the same label for each tile. In particular, Ilse et al.~\cite{ilse2018} propose an attention-based pooling layer aiming to weight each tile individually for its relevance within the bag prediction task. To tackle (ii), recently, convolutional neural networks (CNNs) were combined with self-attention-based Transformers~\cite{Transformer}. Here, the tiles are condensed into feature vectors and subsequently the resulting sequence is fed into a Transformer, where the interdependence between tiles is incorporated by self-attention mechanisms. For instance, Li et al.~\cite{DTMIL} propose a deformable Transformer-based encoder-decoder structure and evaluate it across encoder only based Transformer. Shao et al.~\cite{transmil} uses the Neystr\"om method to approximate self-attention, aiming to decrease the computational complexity. Myronenko et al.~\cite{Myronenko2021} suggest incorporating feature vectors of different scales into an encoder-based Transformer. 

However, general Transformer approaches suffer from quadratic complexity with respect to the sequence length. This complexity is a general problem across computer vision and neural language processing (NLP) domains. To alleviate this concern, Transformer using local attention in the NLP domain~\cite{local_att_nlp} showed that it is sufficient for a token to restrict the attention calculation to a local neighborhood inside the sequence, i.e., the surrounding words. On the other hand, in computer vision self-attention can be modified by introducing local windowed attention~\cite{swin_transformer}. However, this approach for WSIs is not as conveniently applicable as for domain areas where the images have the same size, such as in well-curated datasets like ImageNet~\cite{imagenet}. Whole slide images come with varying geometrical shapes and the representation obtained by using tiles while excluding some entities (e.g. due to artifacts, pen marks, etc) leads to holes within the visual representation. A handcrafted selection of which entities participate in the key-to-query product is not generally applicable for all WSIs, as this would not effectively adapt to the varying local neighborhoods.

Combining the local windowed attention idea with Graph Transformer~\cite{GraphTrans}, we propose a computationally light training pipeline consisting of a CNN and local attention-based Transformer with the following contributions:

\begin{itemize}[noitemsep,topsep=0pt]
  \item We present LA-MIL, a novel local attention graph-based Transformer that restricts self-attention calculations in Transformers by using k-nearest neighbor (kNN) graphs to model local regimes with respect to tiles inside the WSI. 
  \item To our knowledge, LA-MIL is the first pipeline to predict microsatellite instability and tumor mutational burden jointly with genetic alterations as well as the first transformer-based approach for mutation prediction.
  \item An efficiently adapted loss function enables our approach to learn meaningful bag representations for a joint analysis of multiple imbalanced biomarkers.
  \item We evaluate our approach extensively on two datasets for gastrointestinal cancer and demonstrate that our local attention mechanism sufficiently leverages information on par with global self-attention modules. 
  \item LA-MIL shows great modeling interpretability by visualizing local attention scores, consisting of spatial and morphological dependencies. 
\end{itemize}
Our experiments indicate that LA-MIL outperforms state-of-the-art approaches or is on par for mutation prediction tasks in gastrointestinal cancer. 

\section{Method}

\begin{figure}[t]
\centering
\includegraphics[width=\textwidth, keepaspectratio]{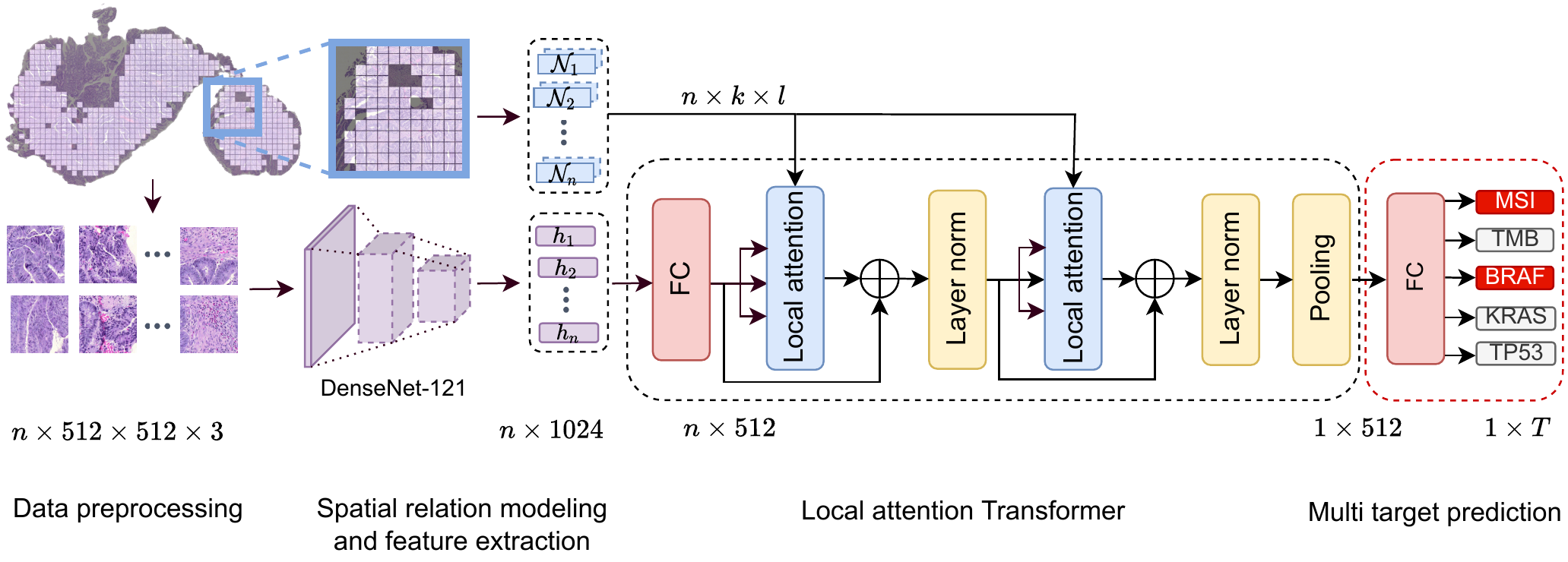}
\caption{LA-MIL overview: The pipeline consists of preprocessing, spatial relation modeling and feature extraction, and a local attention-based Transformer.
} \label{fig:pipeline}
\end{figure}

The pipeline of our LA-MIL approach is visualized in Fig.~\ref{fig:pipeline}. In the following, we introduce the key components of our algorithm. 

\subsection{LA-MIL framework}

\subsubsection{Data preprocessing.}
First, a given gigapixel WSI is tessellated into $N$ tiles, where each tile $t_i \in \mathbb{R}^{H\times W\times C}$. 
Further, we extract the coordinates $c_i \in \mathbb{R}^2$ with respect to the WSI for each tile. Tiles containing background, artifacts, and non tumor-tissue are excluded using Otsu's method~\cite{otsu} and region of interest (RoI) annotations, reducing the number of tiles for downstream processing to $n<N$. 

\subsubsection{Per-tile feature extraction and spatial relation modeling.}
We compress the visual information contained in each tile $t_i$ by extracting features using KimiaNet~\cite{kimianet}, a pretrained DenseNet-121~\cite{densenet}. Thus, the WSI is represented as a sequence of feature vectors $\{h_i\}_{i=1}^{n} \in \mathbb{R}^{n \times D}$, where the dimension $D$ is the output size of the feature extraction CNN. For the spatial relation among the tiles of a WSI, we build $l$ k-nearest neighbor (kNN) graphs $\mathcal{G}_{kNN}^{l}$ using the Euclidean distance of the coordinates $c_i$. A kNN graph can be represented by a matrix $A \in \mathbb{R}^{n \times k}$, indicating the $k$ neighbors for all $n$ tiles.

\subsubsection{Transformer architecture.}
Given a sequence of features $\{h_i\}_{i=1}^{n}$ and $l$ graphs $\mathcal{G}_{kNN}^l$, we further downscale the feature vectors from $D$ to $d$ by using a fully-connected (FC) layer. Subsequently, a Transformer with $l$ blocks of local attention layers is applied. These layers utilize the kNN graphs to update neighboring tiles, thereby modeling local morphological and spatial correlations. Note that a graph can also be shared between layers. Applying a residual connection and layer normalization~\cite{layer_norm} after each attention layer aims to improve the gradient flow and generalization performance. Finally, the sequence is aggregated into a bag-embedding vector $b\in \mathbb{R}^{d}$ by mean pooling as done in~\cite{Transformer}. Another FC layer projects the bag-embedding vector into a target vector $t\in\mathbb{R}^T$, where $T$ is the number of targets to predict. The sigmoid function is applied element-wise on the target vector $t$ to obtain the scores for each target individually. In the context of mutation prediction, the thresholded scores indicate whether a particular gene occurs as wildtype or mutated, respectively. 

\subsubsection{Loss function.}
Mutation prediction is a challenging task since the targets often only occur in small frequencies (see Table~\ref{tab1}). Hence, we use a loss which penalizes the model for wrong decisions about the prediction of underrepresented classes by weighting each binary cross-entropy (BCE) term individually. We take the mean of $T$ BCE losses, thus treating each target equally:
\[
L(\mathbf{x},\mathbf{y}) = - \frac{1}{T} \sum_{t=1}^{T} \frac{n_t^{\text{neg}}}{n_t^{\text{pos}}}\left(y_t \log(\sigma(x_t)) + (1-y_t) \log(1-\sigma(x_t))\right).
\]
As mentioned in the introduction, most mutation prediction studies use tile-level supervised learning rather than bag-level MIL. Hence, a common strategy to tackle highly imbalanced classification is to apply downsampling to reach an equilibrium of classes in the dataset splits~\cite{Kather2020,Kather_msi}. However, this may not be possible in the multi-target and bag-operating setup, as downsampling may exclude nearly all samples, depending on the individual class distributions. 

\subsection{Local attention layer} 
\label{sec:localattention}

\begin{figure}[t]
\centering
\includegraphics[width=\textwidth, keepaspectratio, page=1]{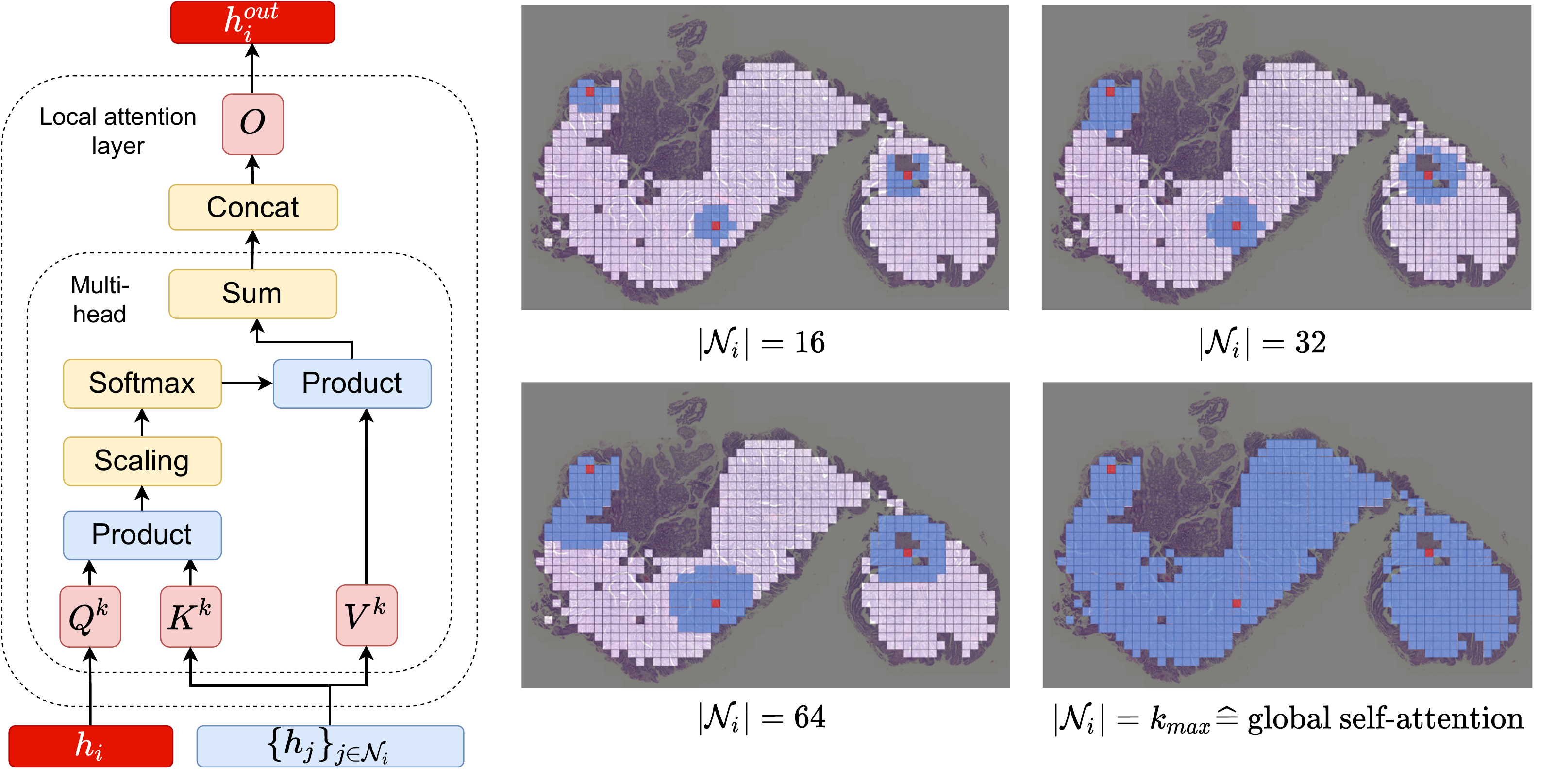}
\caption{Left: Computational block for a local attention layer. Right: Illustration of the spatial field of view for the attention calculations. Tiles in red visualize a query tile and blue colored tiles visualize the adaptive local neighborhoods of different sizes.} \label{fig:local_nbh}
\end{figure}

Self-attention is a key component in Transformer architectures, where each token $h_i$ is updated with global information of the complete input sequence $\{h_1,\ldots, h_{n}\}$. In contrast, our local attention modules constrain the updates for each token $h_i$ associated with node $n_i$ to all tokens $h_j$ with nodes $j\neq i$ that are connected with node $n_i$, as shown in Fig.~\ref{fig:local_nbh}. 
As input we consider a $n$-dimensional sequence of tokens $h_i$ for $i=1,\ldots,n$ associated with $n$ nodes of a graph $\mathcal{G}$ where the nodes are connected by $n\times k$ edges. The update equation for token $h_i$ is
\begin{equation}\label{eq:att_eq}
h_i = O\cdot \mathrm{Concat}_k \left( \sum_{j\in \mathcal{N}_j} w_{ij}^{k} V^k h_{j} \right), \qquad w_{i,j}^{k} = \mathrm{softmax}_j\left(\frac{Q^k h_i \cdot K^k h_j}{\sqrt{d_k}}\right)
\end{equation}
where $Q^k, K^k, V^k \in \mathbb{R}^{d_k \times d}$ and $O \in \mathbb{R}^{d \times d}$ with $k=1,\ldots,H$  denoting the number of the respective attention head. The notation $j\in \mathcal{N}_i$ refers to a set of indices $j$ of nodes connected to the $i$th node by $j$ edges. The cardinality of $\mathcal{N}_i$ is equal to the number of neighbors for all tiles. To calculate local attention scores $a_i$ for each tile from local attention layers, we first cache the intermediate outputs $w_{i,j}^{k}$ from Equation~\ref{eq:att_eq} and sum them across heads and local regimes, i.e.
\begin{equation}\label{eq:att_score_eq}
a_i = \sum_{k=1}^{H}\sum_{j\in \mathcal{N}_j} w_{i,j}^{k}.
\end{equation}
Subsequently, we normalize all $n$ attention values $a_i$ into the range $[0,1]$ and denote the outcomes as local attention scores for each tile.

\section{Experiments}

We applied the proposed method on diagnostic formalin-fixed paraffin-embedded diagnostic slides for two cohorts in the The Cancer Genome Atlas (TCGA) dataset~\cite{TCGA}. From tissue contained in WSIs, we followed recent works~\cite{Fu2020,Kather2020,Coudray2018} and include only tumor-occupied tissue regions. All images were downsampled to 20$\times$ magnification, corresponding to a resolution of $0.5 \frac{\mu m}{\text{px}}$. The task is to predict genetic alterations, the microsatellite status and the tumor mutational burden (TMB) as biomarkers~\cite{Murchan2021} directly from WSIs.

\subsubsection{TCGA colorectal and TCGA stomach datasets.} 
Our dataset TCGA-CRC consists of tiled WSIs from tumor regions of colorectal tissue. We used the preprocessed tumor tissue tiles from \href{https://jnkather.github.io/datasets/}{kather.ai}. As a second dataset, we tiled WSIs of stomach tissue from TCGA, downloaded at \href{https://portal.gdc.cancer.gov}{portal.gdc.cancer.gov}. After noise removal, we excluded all tiles which were not contained in the tumor region by using manual tumor annotations available at: \href{https://jnkather.github.io/datasets/}{kather.ai}. We retrieved the genetic annotations matching the WSIs from \href{https://xenabrowser.net/datapages/}{xenabrowser.net}. Annotations for the microsatellite stability/instability (MSS/MSI) and TMB are available at \href{https://cbioportal.org}{cbioportal.org}. Following Luchini et al.~\cite{Luchini2019}, we binarized MSS and MSI-Low as MSS and MSI-High as MSI. 

\begin{table}[t]
\caption{Distribution of genetic alterations in TCGA-CRC and TCGA-STAD. We denote the number of positive samples for each target.}\label{tab1}
\begin{tabular}{c|c|cccccccccc}
\hline
Cohort & n   & MSI & TMB & BRAF & ALK & ERBB4 & FBXW7 & KRAS & PIK3CA & SMAD4 & TP53 \\ \hline \hline
CRC     & 594 & 78  & 85  & 66   & 40  & 62  & 107    & 223  & 178    & 81   & 332    \\ 
STAD    & 440 & 75  & 86  & 13   & 19   & 62  & 38     & 40  & 86    & 36    & 225    \\ \hline
\end{tabular}
\end{table}

\subsubsection{Implementation.}
Each tile was embedded into a 1024-dimensional feature space by a DenseNet-121 model that was pretrained on histopathological data. By using the coordinates of each tile, we built two kNN-Graphs with $k=16$ and $k=64$ for subsequent attention restriction in the first and second local attention module, respectively. In the training phase, each feature vector associated with the tiles was further compressed from 1024 to 512 by a FC layer. After a stack of two local attention layers, we averaged the feature vectors across all tiles. The resulting bag embedding was passed through a classification head, consisting of another fully connected layer from 512 to 10, to compute the logits. We applied the sigmoid activation function element-wise to calculate the probabilities for each individual target.  For optimization, we employed the Lookahead optimizer~\cite{lookahead} together with AdamW~\cite{adamw}, and used a learning rate of 2e-05 and 2e-04 (for TCGA-CRC and TCGA-STAD, respectively) for 10 epochs, weight decay of 2e-05, and batch size 1. The LA-MIL model with 2.1M parameter was implemented in PyTorch and DGL~\cite{dgl_lib}, and trained on a single Tesla V100 GPU. 

\subsubsection{Evaluation.}
To evaluate the mutation prediction task on both datasets TCGA-CRC and TCGA-STAD, we compared the performance of LA-MIL with state-of-the-art methods. The fact that most of the recent advances predict $T$ biomarkers individually results in training, validating, and hyperparameter tuning for $T$ separate models, while we used a single model to predict all biomarkers. Moreover, we implemented a Transformer MIL approach where we exchanged all local attention blocks with global self-attention~\cite{Transformer}, denoted as T-MIL. To stick with common evaluation procedures for mutation prediction in recent works, we evaluated our pipeline with a 5-fold cross validation (CV). We split the datasets into folds such that individual class distributions in each fold were approximately the same and ensured that no patient appeared in the training and validation set at the same time. We measured the performance using the area under the receiver operating characteristic curve (AUROC) for each target individually. 

\section{Results}

\begin{table}[t]
\caption{Mean AUROC scores for mutation prediction on the datasets TCGA-CRC and TCGA-STAD. For the competitive methods, we report results from the original publications; for our methods, we report the mean over five folds (see supplementary material for results with standard deviation).}\label{tab:results_crc}
\centering
\begin{tabular}{c|l|cccccccccc}
\hline
\rotatebox{90}{Dataset} & Method                  & \rotatebox{90}{MSI}   & \rotatebox{90}{TMB}   & \rotatebox{90}{BRAF}  & \rotatebox{90}{ALK}   & \rotatebox{90}{ERBB4} & \rotatebox{90}{FBXW7} & \rotatebox{90}{KRAS}  & \rotatebox{90}{PIK3CA} & \rotatebox{90}{SMAD4} & \rotatebox{90}{TP53}  \\ \hline \hline
\multirow{6}{*}{\rotatebox{90}{TCGA-CRC}}& Kather et al.~\cite{Kather_msi}
                            & 0.77          & --            & --            & --            & --            & --            & --            & --            & --            & --            \\
& Wang et al.~\cite{Wang2020}      & --            & 0.82          & --            & --            & --            & --            & --            & --            & --            & --            \\
& Kather et al.~\cite{Kather2020}  & --            & --            & 0.66          & 0.51          & --            & 0.49          & 0.60          & \textbf{0.62} & \textbf{0.63} & \textbf{0.68} \\
& Fu et al.~\cite{Fu2020}          & --            & --            & 0.57          & --            & --            & \textbf{0.66} & 0.55          & 0.59          & 0.58          & \textbf{0.68} \\
& T-MIL (Ours)              & \textbf{0.85} & 0.82          & \textbf{0.73} & 0.61          & 0.57          & 0.64          & 0.61          & 0.60          & 0.60          & 0.64 \\
& LA-MIL (Ours)             & \textbf{0.85} & \textbf{0.83} & 0.72          & \textbf{0.63} & \textbf{0.60} & \textbf{0.66} & \textbf{0.62} & 0.61          & 0.58          & 0.63 \\ \hline \hline

\multirow{6}{*}{\rotatebox{90}{TCGA-STAD}} & Kather et al.~\cite{Kather_msi}
                            & \textbf{0.81}  & --  & --  & --  & --  & --  & --  & --  & --  & --  \\
& Wang et al.~\cite{Wang2020}      & --  & 0.75  & --  & --  & --  & --  & --  & --  & --  & --  \\
& Kather et al.~\cite{Kather2020}  & --  & --   & 0.37  & 0.45  & --   & \textbf{0.74}  & 0.64  & \textbf{0.67}  & 0.61  & 0.60  \\
& Fu et al.~\cite{Fu2020}          & --   & --   & --  & --   & --   & --   & --   & 0.47  & 0.49  & \textbf{0.63}  \\ 
& T-MIL (Ours)              & 0.80  & \textbf{0.78}  & \textbf{0.73}  & \textbf{0.52}  & \textbf{0.47}  & 0.71  & 0.65  & 0.58  & 0.62  & 0.57  \\
& LA-MIL (Ours)             & 0.78  & 0.77  & 0.67  & \textbf{0.52}  & \textbf{0.47}  & 0.72  & \textbf{0.70}  & 0.61  & \textbf{0.64}  & 0.58  \\ \hline
\end{tabular}
\end{table}
Current methods, such as Kather et al.~\cite{Kather_msi} for MSI and Wang et al.~\cite{Wang2020} for TMB, predict the biomarkers instance-wise as single targets for each tile with the corresponding label inherited from its parent WSI. Similarly, Kather et al.~\cite{Kather2020} and Fu et al.~\cite{Fu2020} train one model for each target gene when predicting mutations, but evaluate their results on WSI-level by average pooling of tile-wise predictions. In contrast, we train and evaluate only one model on WSI-level to predict multiple biomarkers using a MIL transformer that aggregates features from all tiles with global or local self-attention layers. 

Table~\ref{tab:results_crc} shows the AUROC scores of state-of-the-art instance-wise methods compared to our methods on the dataset TCGA-CRC. The results suggest that our models can leverage information from multiple targets to achieve better overall performance. Interestingly, this holds especially for the prediction of MSI, where our approach improves the score by 8\% from 0.77 to 0.85. The prediction of TMB and most of the gene mutations (except for SMAD4 and TP53) are on par or marginally better (+1\%).

The results for the TCGA-STAD dataset in MSI prediction are marginally worse (-1\%). This could arise due to the fact that the compared work uses a slightly different label strategy for MSI-Low cases, which also affects the evaluation. The results for TMB improves by up to 3\%. Similar as on the TCGA-CRC dataset, we observe an improvement of results using our Transformer-based approaches for the remaining targets except for a few genes.

\subsubsection{Local attention visualization.}
\begin{figure}[t] 
\centering
\includegraphics[width=\textwidth, keepaspectratio]{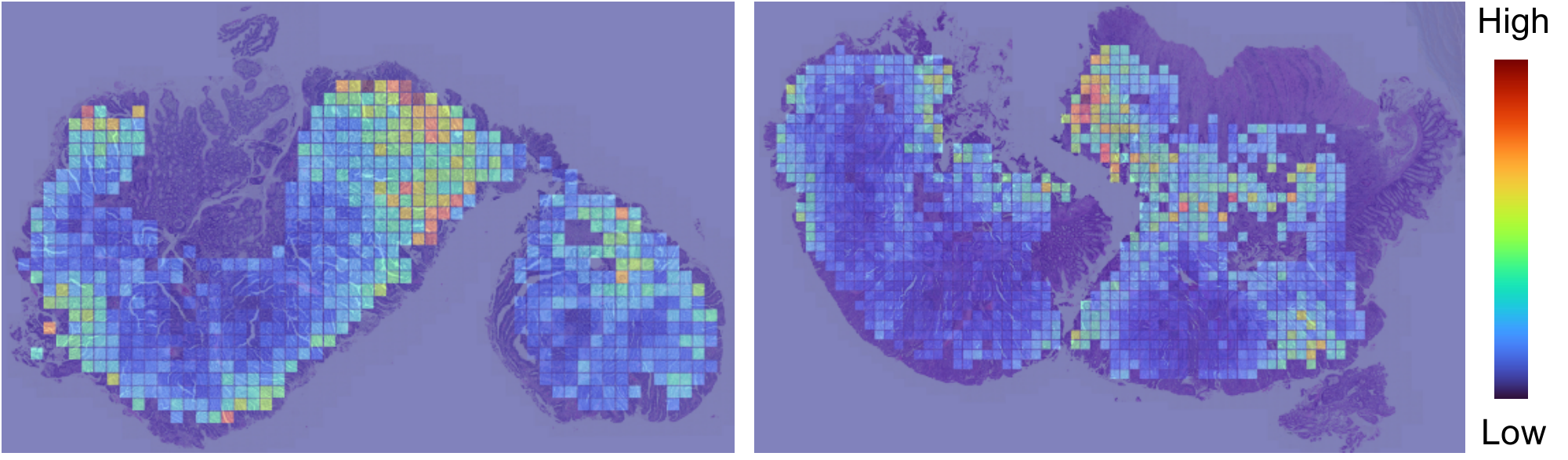}
\caption{Local attention scores visualization for the last local attention layer with restricted self-attention in a neighborhood of size 64.}\label{fig:local_att}
\end{figure}

As described in Eq. \ref{eq:att_score_eq}, we can calculate the local attention scores from the query-to-key product. In Fig. \ref{fig:local_att} we colorized the tiles according to their corresponding local attention score. The high attentive regions include the nuclear chromatin that seems to be hyperchromatic, as well as crowded glands or solid areas. Yet, to the best of our knowledge, there are no distinguishing patterns from WSI for mutated genes, which makes it difficult for a quantitative evaluation. Nevertheless, LA-MIL provides an interpretation based on the contribution of each tile for the bag-level prediction task and thus paves a way for a deeper investigation of highly scored tiles. 

\section{Conclusion}
In this work, we proposed a novel MIL framework with local attention for WSI analysis. Local attention is achieved through a graph-based transformer that models region-wise inter-dependencies. As the size of the region can be set arbitrarily, our approach bridges the gap between instance-wise and global relation approaches by modeling local relations of arbitrary size. Moreover, an effective adapted loss enables us to learn multiple biomarkers at once, for low computational cost compared to CNN-only based methods.

However, there is often more than one WSI for a patient available in the TCGA database. Future work will investigate the strategy of combining all WSIs for a patient, while suitably scaling the coordinates for tiles of different WSIs. Thus, each tile will only be updated with local information from its direct parent WSI and transformed into a bag-embedding consisting of locally correlated tiles of all WSIs from a patient. We believe that our approach provides a base for further applications in other WSI analysis tasks where structured relation modeling is crucial.

\subsubsection{Acknowledgements} 
S.J.W. was supported by the Helmholtz Association under the joint research school ''Munich School for Data Science - MUDS''.

\bibliographystyle{splncs04}
\bibliography{bibliography}

\end{document}